# Dual Signal Decomposition of Stochastic Time Series

**Alex Glushkovsky**

BMO Financial Group

## Abstract

The decomposition of a stochastic time series into three component series representing a dual signal—namely, the mean and dispersion—while isolating noise is presented. The decomposition is performed by applying machine learning techniques to fit the dual signal. Machine learning minimizes the loss function which compromises between fitting the original time series and penalizing irregularities of the dual signal. The latter includes terms based on the first and second order derivatives along time. To preserve special patterns, weighting of the regularization components of the loss function has been introduced based on Statistical Process Control methodology. The proposed decomposition can be applied as a smoothing algorithm against the mean and dispersion of the time series. By isolating noise, the proposed decomposition can be seen as a denoising algorithm. Two approaches of the learning process have been considered: sequential and jointly. The former approach learns the mean signal first and then dispersion. The latter approach fits the dual signal jointly. Jointly learning can uncover complex relationships for the time series with heteroskedasticity. Learning has been set by solving the direct non-linear unconstrained optimization problem or by applying neural networks that have sequential or twin output architectures. Tuning of the loss function hyperparameters focuses on the isolated noise to be a stationary stochastic process without autocorrelation properties. Depending on the applications, the hyperparameters of the learning can be tuned towards either the discrete states by stepped signal or smoothed series. The decomposed dual signal can be represented on the 2D space and used to learn inherent structures, to forecast both mean and dispersion, or to analyze cross effects in case of multiple time series.

## 1 Introduction

Time series analysis is a fundamental topic in statistical and machine learning research, with a rich history encompassing seminal works such as the influential book "Time Series Analysis: Forecasting and Control" (Box and Jenkins, 1970), numerous journal articles, dedicated academic journals, and specialized conferences. This field addresses key concepts including seasonality, trends, autocorrelation, Autoregressive Integrated Moving Average (e.g., ARIMA models), heteroskedasticity (e.g., GARCH models), multivariate analysis, and modern deep learning techniques such as recurrent neural networks (RNNs) and long short-term memory (LSTM) networks. The most recent approaches include Transformer-based models, such as constrained reinforced learning (Liu and Liu, 2021), long-term time series forecasting (LTSF) (Zeng *et al*, 2022), and multivariate time series forecasting (Nie *et al*, 2022). Given the vast scope of time series methodologies, this article does not aim to provide a comprehensive overview.

Most approaches to time series modeling can be broadly classified into two categories: (1) models in which the current value of the time series is expressed as a function of past values, excluding the current value, with additional residual terms representing an error component (e.g., predictive models for forecasting future observations), and (2) models in which the current value is expressed as a function that includes the present observation itself, allowing for decomposition and structural analysis. The first category is primarily used for forecasting, while the second facilitates decomposition and interpretability of time series components.

This article addresses the second category, where the time series $X_t$, $t \in [1, T]$ is decomposed into a dual signal, i.e., the mean $M_t$ and dispersion $S_t$, and noise $\varepsilon_t$ into the following simple form:

$$X_t = M_t + S_t \varepsilon_t, \; S_t \geq 0 \qquad (1)$$

This form is similar to standardized variable transformation but instead of constant statistics, there are learned time series of the dual signal $M_t$ and $S_t$.

Equation (1) represents a generalized process of time series (Brockwell and Davis, 2009; Dagum, 2010; Bernardi *et al*, 2024) that in addition to a single mean signal $M_t$ addresses the heteroscedasticity by dispersion signal $S_t$. The second term in equation (1) that splits the error term into stochastic stationary noise $\varepsilon_t$ and time dependant dispersion is a key element of the ARCH process (Elad, 1982). Extractions of dispersion along with trend and seasonality components of the time series have been addressed in a seasonal-trend-dispersion decomposition (STD) approach (Dudek, 2023). The dispersion component in STD is represented as an interaction with seasonality effects in addition to the trend, while form (1) includes the dispersion as an interaction with the noise.

Assumably the decomposed dual signal represents a more systematic, structured, and, therefore, a potentially better prediction of the time series. In addition, it represents a denoising approach with abilities to discover inherent dependence between dual signals. The dual signal decomposition (1) supports further decomposition of the signals into seasonal (cycling), shifting, and trending series, and the identification of outliers.

## 2    Dual signal decomposition

The decomposition (1) of the stochastic time series $X_t$ by dual signal and noise is based on the following guiding principles:

1. To identify a dual signal concerning the mean $M_t$ and the dispersion $S_t$ (i.e., heteroskedasticity), while isolating noise $\varepsilon_t$.
2. To preserve special patterns of the time series, such as outliers, trends, shifts, or cycling.
3. To isolate noise that should be a stationary time series whose statistical properties (mean, dispersion, and autocorrelation) remain about constant over time, while mean and autocorrelation should be close to zero. Ideally isolated noise $\varepsilon_t$ should be a white noise.
4. To control the signal's smoothness across a wide range, from step functions to continuous signals, by enforcing regularization and smoothness.
5. To sequentially or jointly learn the mean and dispersion (heteroskedasticity) signals.

The learning of the dual signals can be applied sequentially or jointly. The former approach first independently fits time series $X_t$ by the mean signal $M_t$ and then learns dispersion $S_t$ based on sign-invariant residuals $|R_t| = |X_t - M_t|$. The jointly learning approach fits the dual signal simultaneously.

Let us consider sequential learning first. Any machine learning approach is guided by a loss function, which serves as an objective to be minimized. The proposed loss function to learn the mean signal $M_t$ given the time series $X_t$ as an input has three components, i.e., fitting, regularization, and smoothing:

$$Loss_M(X_t) = arg\,min_{M_t}(Fitting(X_t, M_t) + \\ \beta_M \cdot Regularization(M_t, M_{t-1}) + \\ \gamma_M \cdot Smoothing(M_t, M_{t-1}, M_{t-2})) \qquad (2)$$

where hyperparameters $\beta_M$ and $\gamma_M$ are Lagrangian multipliers controlling the contribution of the regularization and the smoothing components, correspondingly.

The fitting component can be calculated using various expressions of sign-invariant deviations $|X_t - M_t|$ pushing the mean time series $M_t$ to be close to the original time series $X_t$. The most common metrics are:

- Root Mean Squared Error: $RMSE(X_t, M_t) = \sqrt{E[(X_t - M_t)^2]} = \sqrt{\frac{1}{T}\sum_{t=1}^{T}(X_t - M_t)^2}$
- Mean Squared Error: $MSE(X_t, M_t) = E[(X_t - M_t)^2] = \frac{1}{T}\sum_{t=1}^{T}(X_t - M_t)^2$
- Mean Absolute Error: $MAE(X_t, M_t) = E[|X_t - M_t|] = \frac{1}{T}\sum_{t=1}^{T}|X_t - M_t|$
- Sum of Squared Error: $SSE(X_t, M_t) = \sum_{t=1}^{T}(X_t - M_t)^2$
- Maximum of Squared Error: $MAXSE(X_t, M_t) = max_t((X_t - M_t)^2)$
- Maximum of Absolute Error: $MAXAE(X_t, M_t) = max_t(|X_t - M_t|)$

The first three expressions are normalized by the number of observations $T$, unlike the Sum of Squared Error, which reflects the total (unnormalized) error, and the last two approaches, which capture the worst-case errors.

This paper is limited by using Root Mean Square Error (RMSE) as a fitting metric.

The regularization component can be formalized against the sign-invariant changes of the sequential values $|M_t - M_{t-1}|$. This term introduces penalties to changes of the mean time series along time. Similarly to the fitting component, it can be done by using different approaches from a simple mean of the weighted moving ranges to more complicated rolling deviations or variances.

This paper addresses the following two versions of regularization:

- Mean weighted moving range (MAE version):

$$Regularization(M_t, M_{t-1}) = E[w_t \cdot |\Delta M_t|] = \frac{1}{T-1}\sum_{t=2}^{T} w_t \cdot MR(M_t)$$

- Root Mean Squared weighted moving range (RMSE version):

$$Regularization(M_t, M_{t-1}) = \sqrt{E[w_t \cdot (M_t - M_{t-1})^2]} = \sqrt{\frac{1}{T-1}\sum_{t=2}^{T} w_t \cdot (M_t - M_{t-1})^2}$$

where $w_t \in [0, 1]$ are weights preserving special patterns of the time series $X_t$, and $MR_t$ are moving ranges $MR_t = |\Delta M_t| = |M_t - M_{t-1}|$, $t \geq 2$.

The regularization terms presented above are commonly referred to as L1 and L2 norms, correspondingly. They influence model behavior differently: L1 regularization often leads to sparse, step-like solutions, while L2 regularization produces smoother, more continuous outputs.

A pairwise compatibility chart between fitting and regularization components using different expressions is shown in Table 1. It takes into consideration the order and scale (normalization) of the underlying metrics. For example, MAE is a first order metric that normalized by the number of observations *T*. By contrast, SSE is a second order and unnormalized.

|       | RMSE    | MSE     | MAE     | SSE     | MAXSE   | MAXAE |
|-------|---------|---------|---------|---------|---------|-------|
| RMSE  |         |         |         |         |         |       |
| MSE   | ≠O; ≠N  |         |         |         |         |       |
| MAE   | ✓O; ≠N  | ≠O; ✓N  |         |         |         |       |
| SSE   | ≠O; ≠N  | ✓O; ≠N  | ≠O; ≠N  |         |         |       |
| MAXSE | ≠O; ✓N  | ✓O; ✓N  | ≠O; ✓N  | ✓O; ≠N  |         |       |
| MAXAE | ✓O; ✓N  | ≠O; ✓N  | ✓O; ✓N  | ≠O; ≠N  | ≠O; ✓N  |       |

Legend:
✓ compatible
≠ incompatible

Table 1. Pairwise compatibility between metrics based on order (O) and normalization (N)

The smoothing component is defined as the weighted mean of the absolute values of the second order derivatives along time:

$$Smoothing(M_t, M_{t-1}, M_{t-2}) = E[w_t \cdot |\Delta^2 M_t|] = \frac{1}{T-2} \sum_{t=3}^{T} w_t \cdot MR(\Delta M_t)$$

where $\Delta^2 M_t = MR(\Delta M_t) = M_t - 2 \cdot M_{t-1} + M_{t-2}$.

While this article is restricted to the use of MAE for the smoothing component, alternative evaluation metrics may also be applicable.

Applying RMSE for the fitting component and MAE for both regularization and smoothing components, the loss function can be defined as:

$$Loss_M(X_t) = arg\ min_{M_t}(RMSE(X_t, M_t) + \beta_M \cdot E[w_t \cdot |\Delta M_t|] + \gamma_M \cdot E[w_t \cdot |\Delta^2 M_t|]) \quad (3)$$

While there is some overlap between regularization and smoothing, incorporating both provides greater flexibility for capturing first- and second-order variability in the mean signal time series, as each can be tuned independently through the loss function's hyperparameters.

The form (3) can be interpreted as a modified version of the Potts or the Blake-Zisserman functionals (Blake and Zisserman, 1987; Mumford and Shah, 1989; Winkler *et al*, 2003; Elad, 2010; Weinmann *et al*, 2012) or generalized functional denoising piecewise constant signals (Little and Jones, 2011).

The Potts functional has both fitting and regularization terms. The former term is a sum of squared deviations (i.e., SSE) between the mean and time series. The later term minimizes the number of mean's jumps where $M_{t-1} \neq M_t$. It leads to a jump-sparse approximation of the time series by the single $M_t$ signal.

The proposed modification includes the following improvements compared to the Potts functional:

- Regularization component is either a mean of the weighted moving ranges or the root mean squared weighted moving ranges. Even the moving ranges are smoother than the conditional $M_{t-1} \neq M_t$ counts. It supports an easier optimization process that minimizes the loss function.

Therefore, both applied options provide more informative and flexible approaches compared to the number of jumps.

- Introduced weightings of the regularization and smoothing components allow for preservation of special patterns of the time series $X_t$. The impact of the weightings can be controlled by the hyperparameters.
- The decomposition addresses the dual signal, i.e., not only the mean time series but heteroscedasticity as well. The learning of the dual signal can be done sequentially or jointly.
- The added smoothing component controls instability of the second order deviations of the mean time series (i.e., $\Delta^2 M_t$).

To provide some easily interpretable insights into the learning process, let us consider a simple example of sequential learning that includes only the fitting and regularization components. For simplicity, both components are defined based on the mean squared error (MSE) metric, and the regularization component does not preserve any special patterns—that is, the weights $w_t$ are equal to one ($w_t \equiv 1$):

$$Loss_M(X_t) = arg\ min_{M_t}(MSE(X_t, M_t) + \beta_M \cdot MSE(\Delta M_t))$$

The latest can be simplified as

$$Loss_M(X_t) \sim arg\ min_{M_t}((\bar{X} - \bar{M})^2 - 2 \cdot \rho_{XM} \cdot \sigma_X \cdot \sigma_M + (1 + 2 \cdot \beta_M \cdot (1 - r)) \cdot \sigma_M^2) \quad (4)$$

where $\bar{X}$ and $\bar{M}$ denote the means of the time series $X_t$ and $M_t$, respectively; $\sigma_X$ and $\sigma_M$ are their standard deviations; $\rho_{XM}$ is the cross-correlation coefficient between time series $X_t$ and $M_t$; and $r$ represents the autocorrelation coefficient of the time series $M_t$.

The simplified form of the loss function (4) implies the following conditions for its minimization:
- $\bar{M} \to \bar{X}$
- $\rho_{XM} \to 1$
- $\sigma_M \to 0$
- $r \to 1$

The first two conditions encourage the mean signal $M_t$ to closely match the mean and temporal alignment of the original series $X_t$, while the latter two aim to suppress volatility in $M_t$ and promote its temporal consistency through higher autocorrelation.

The Lagrangian multiplier, i.e., the hyperparameter $\beta_M$, controls the contribution of the latter two conditions.

It is worth noting that the obtained results are also valid for the sum of squared error (SSE) losses of both fitting and regularization components.

To address the second signal $S_t$, i.e, the dispersion of the residual time series $R_t = X_t - M_t$, different sign-invariant functions can be used, such as absolute values $AE = |R_t|$ or squared error $SE = R_t^2$. The article is limited to the former option of the dispersion component, while the regularization and smoothing components have a similar structure to the first stage of the mean decomposition (3):

$$Loss_S(R_t) = arg\ min_{S_t}(RMSE(|R_t|, S_t) + \beta_S \cdot \boldsymbol{E}[w_t \cdot |\Delta S_t|] + \gamma_S \cdot \boldsymbol{E}[w_t \cdot |\Delta^2 S_t|]) \quad (5)$$

Depending on the application, the weighting $w_t$ of the regularization and smoothing components in the dispersion loss function (5) can be dropped or kept.

## 2.1 Dual signal

The dual signal decomposition (Equation 1) addresses both mean and dispersion while constraining the isolated noise to be stationary.

The dual signal approach is a valuable technique for capturing different stochastic outcomes. It offers a more comprehensive view by simultaneously considering the expected behavior (mean) and the variability around it (dispersion). While the most well-known methodologies in time series analysis—such as ARCH, GARCH, and their extensions—focus on modeling both mean and variance, the dual signal approach extends beyond time series. It has also been effectively applied in the analysis of designed experiments (DOE) and regression models through a method known as dual response. Notably, non-linear optimization methods that combine Response Surface Methodology (RSM) with dual response techniques have been explored in works such as Vining and Myers (1990), Castillo and Montgomery (1993), and Lin and Tu (1995). The dual response framework has proven useful for the robust tuning of machine learning models (Glushkovsky, 2018).

## 2.2 Preservation of special patterns

The regularization and smoothness terms of the loss function can mask special patterns of the time series. To preserve special patterns, the Statistical Process Control (SPC) methodology has been employed to distinguish between common cause variation ("in-control" state) and assignable (i.e., special) cause variation ("out-of-control" state) (Juran and De Feo, 2016; Montgomery, 2020).

Common SPC for individual data points (IX) relies on ±3 sigma control limits, defined as follows:
Lower Control Limit: $LCL = \bar{X} - 3 \cdot \overline{MR}/d_2$
Upper Control Limit: $UCL = \bar{X} + 3 \cdot \overline{MR}/d_2$
where $\bar{X}$ is the process mean, $\overline{MR}$ is the average moving range, and $d_2 = 1.128$ is a constant used to convert the average moving range into an estimate of the standard deviation.

Data points that are closer to the control limits have a higher likelihood of being influenced by assignable causes rather than common variation. Consequently, such points may indicate deviations from the expected process behavior rather than natural process variation.

By adapting SPC methodology and assuming that neighboring data points are normally distributed, Z-values for time series observations can be estimated using rolling averages of the process mean ($\bar{X}$) and moving range ($\overline{MR}$) calculated from either the preceding or succeeding *n* points. The absolute Z-value at time *t* is defined as:

$$|Z_t| = \frac{d_2 \cdot |X_t - MAn_t|}{MAn_t(MR_t)}, Z_t \in [0, +\infty)$$

where:

- $MAn_t$ is the moving average of the time series, computed from either the preceding *n* points $MAn_t^p = \frac{1}{n}\sum_{\tau=t-n}^{t-1} X_\tau$ or the succeeding *n* points $MAn_t^s = \frac{1}{n}\sum_{\tau=t+1}^{t+n} X_\tau$
- $MR_t = |X_t - X_{t-1}|$ is the moving range at time *t*

This formulation provides a standardized measure of how much a time point deviates from its local context, accounting for local variability.

Depending on application, estimations of absolute Z-values based on both preceding and succeeding calculations can be aggregated as $|Z_t| = \max(|Z_t^p|, |Z_t^s|)$ or just $|Z_t| = |Z_t^p|$. The former option is suitable for analyzing historical time series, while the latter is a backward-looking approach better suited for dynamically evolving time series.

Typically, "out-of-control" points are identified when the absolute Z-value of the current observation $X_t$ exceeds 3.

The illustration of the control limits calculated based on the above formulas applying preceding $n$ = 7 points as a control window is shown in Figure 1. It can be observed that the point $t$ = 40 is in an "in control" state, while point $t$ = 50 is in an "out-of-control" state and represents an outlier.

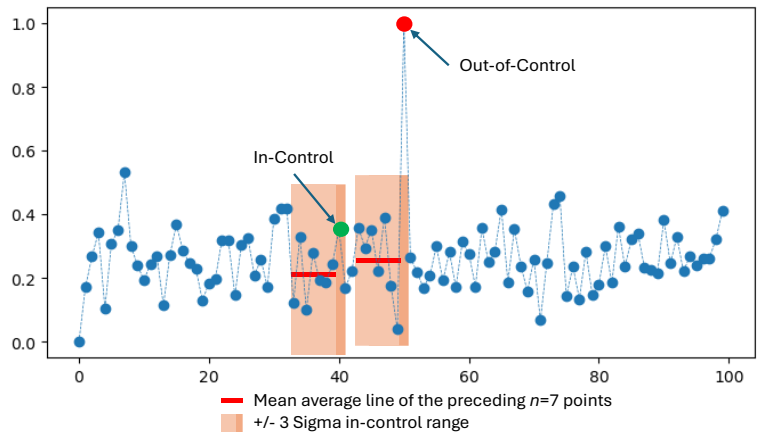

Figure 1. Illustration of the Statistical Process Control chart for individual data points (IX)

The calculated absolute Z-values can then be converted into $p$-values:

$$p_t = 2 \cdot (1 - \Phi(|Z_t|)), p_t \in (0, 1]$$

where $\Phi(\cdot)$ is the cumulative distribution function (CDF) of the standard normal distribution.

The lower the $p$-value, the more likely it is that the point has been influenced by assignable causes, indicating that the signal should closely match the observed value in the time series to preserve special patterns. Therefore, the contribution of the regularization component in the loss function should be reduced for such points.

To flexibly preserve the special patterns, different weights for the regularization component of the loss function have been considered based on the absolute Z-value and the calculated $p$-value, consequently (Figure 2).

The choice of the weighting function's dependence on the $p$-value can be incorporated into the tuning process to enforce the stationarity property of the noise.

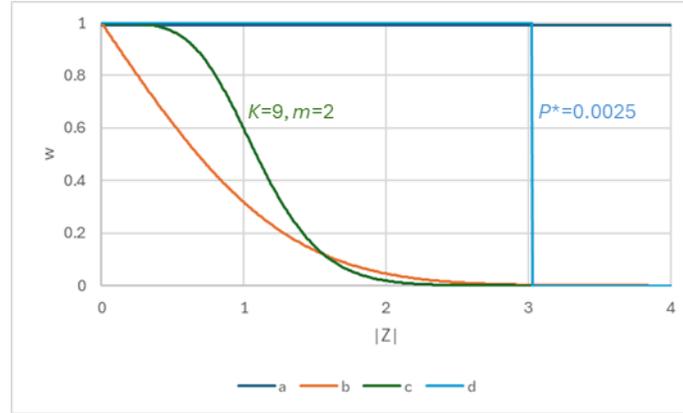

Figure 2. Weights of the regularization component versus the absolute Z-value:
- a) no weighting: $w_t \equiv 1$ for any $t \in [1, T]$
- b) linear: $w_t = p_t$
- c) transformed: $w_t = 1 - \exp(-k \cdot p_t^m)$,
  where $k$ and $m$ are hyperparameters, such as $k = 9$ and $m = 2$
- d) binary: $w_t = 0$, if $p_t \leq p^*$ or 1, otherwise,
  where $p^*$ is a cutoff value, such as 0.0025

While the Individual IX SPC approach handles outliers reasonably well, more advanced SPC techniques—such as incorporating run tests (also known as Western Electric rules)—can be applied to better detect and preserve distinctive trends, shifts, or cycles in the decomposed time series (Juran and De Feo, 2016; Montgomery, 2020). Alternatively, it can be set by applying SPC techniques that value neighbor's distances, for example, based on exponentially weighted moving average (EWMA) or utilizing process stability index (PSI) instead of *p*-values (Glushkovsky, 2006).

The rationale for preserving special patterns of the time series $X_t$ can be illustrated by the three examples shown in Figure 3, which respectively demonstrate: an outlier (a, b), a shift in the mean (c, d), and a cyclical pattern (e, f).

The decompositions of the simulated time series in Figure 3 are based on the following settings: Fitting – RMSE, Regularization – MAE, $\beta_M = \beta_S = 3.9$ and have been calculated using Equation (6), $\gamma = 0.5$, and $\theta = 0$.

It turned out that the introduced binary weightings (charts on the right side) support more adequate decompositions while preserving special patterns. The decompositions without weighting tend to be over smoothed (charts on the left side).

The examples in Figure 3 have been simulated with known introduced noises. The following correlations between introduced simulated and decomposed noises clearly confirm the improvements after applying binary weightings towards special patterns preservations: 98% vs 95%, 98% vs 89%, and 77% vs 62% considering charts with outlier, mean shift and cycling, correspondingly.

It can be observed that the binary weightings $w_t$ mostly equal 1 except for periods of significant changes when the special patterns are present (Figure 3, right side). During such periods, the regularization component is switched out by weights equal to zero.

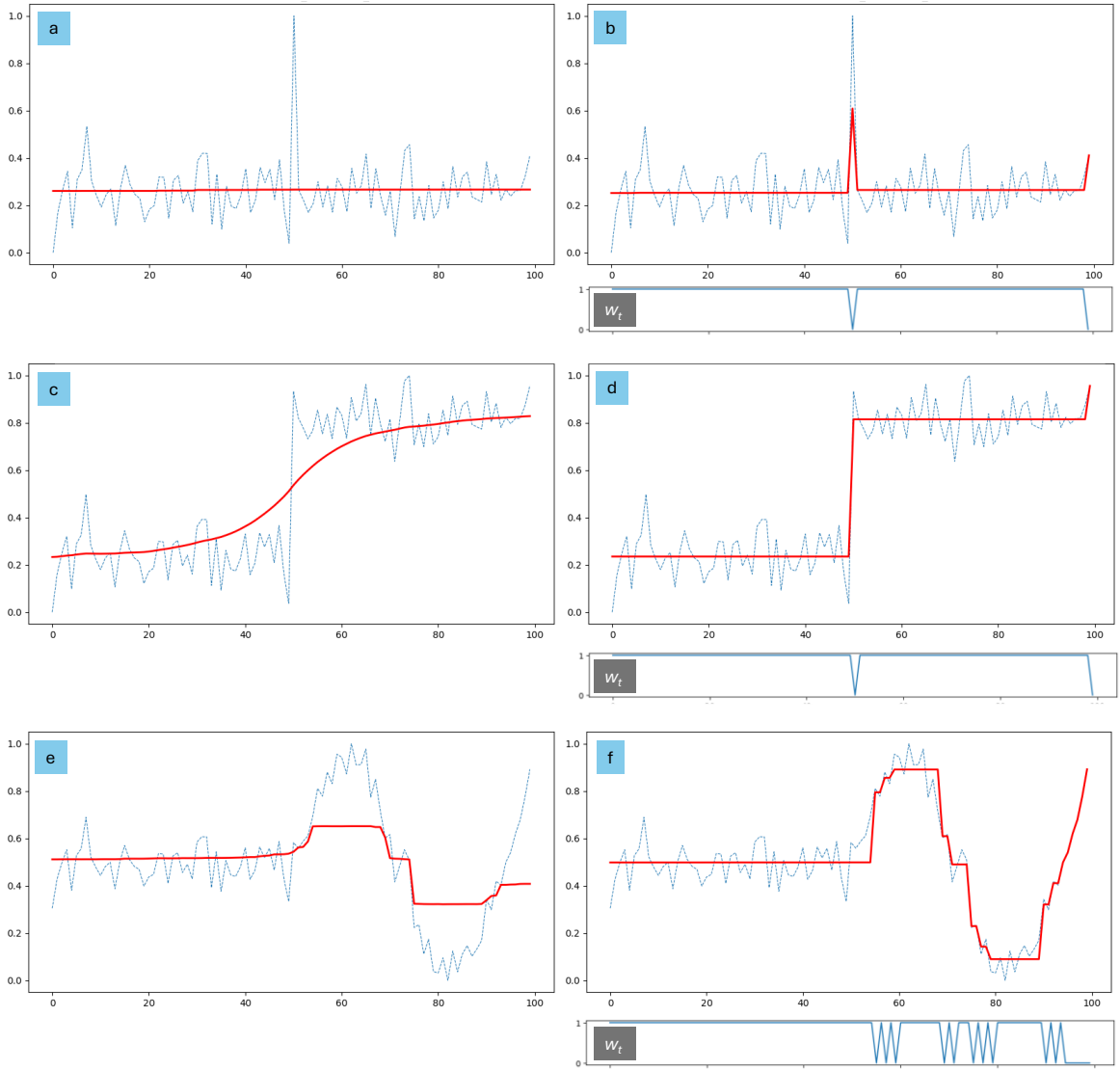

Figure 3. Illustrations of the weighting effects on the decomposed mean time series $M_t$:
- the left charts (a), (c), and (e) are decompositions without weighting ($w_t \equiv 1$)
- the right charts (b), (d), and (f) are decompositions with binary weightings ($w_t = 0$, if $p_t \leq 0.0025$ or 1, otherwise).

## 2.3 Stationarity of the isolated noise

According to Equation (1), noise can be isolated as $\varepsilon_t = (X_t - M_t)/S_t$ and it should be at least a weak stationary time series whose statistical properties (mean, dispersion, and autocorrelation) remain about constant over time, while mean and autocorrelation should be close to zero. Ideally isolated noise $\varepsilon_t$ should be a white noise.

Stationarity of the noise can be assessed by applying one or more of the following tests along with their corresponding metrics:

- <u>Augmented Dickey-Fuller (ADF) test</u>: Assesses stationarity by checking for unit roots in the $\varepsilon_t$ time series (Dickey and Fuller, 1979).
- <u>Ljung-Box (LB) test</u>: Evaluates whether the $\varepsilon_t$ time series exhibit autocorrelation, ensuring the noise behaves like white noise (Ljung and Box, 1978).
- <u>Rolling statistics</u>: Examines the stability of the rolling mean and standard deviation over time.

Moreover, the minimization of the Kullback–Leibler (K–L) divergence can be employed to ensure that the distribution of the isolated noise closely aligns with the standard normal distribution *N*(0,1) (Kullback and Leibler, 1951).

Depending on the application, additional tests can be introduced by enforcing independence between the isolated noise and learned dual signal.

Stationary statistics of the noise can be used to tune hyperparameters of the loss function. It means applying tuning that enforces noise stationarity (Figure 4). Enforcement of the noise towards a stationary time series is a widely used approach, for example (Livieris *et al*, 2021; Zhong, 2023).

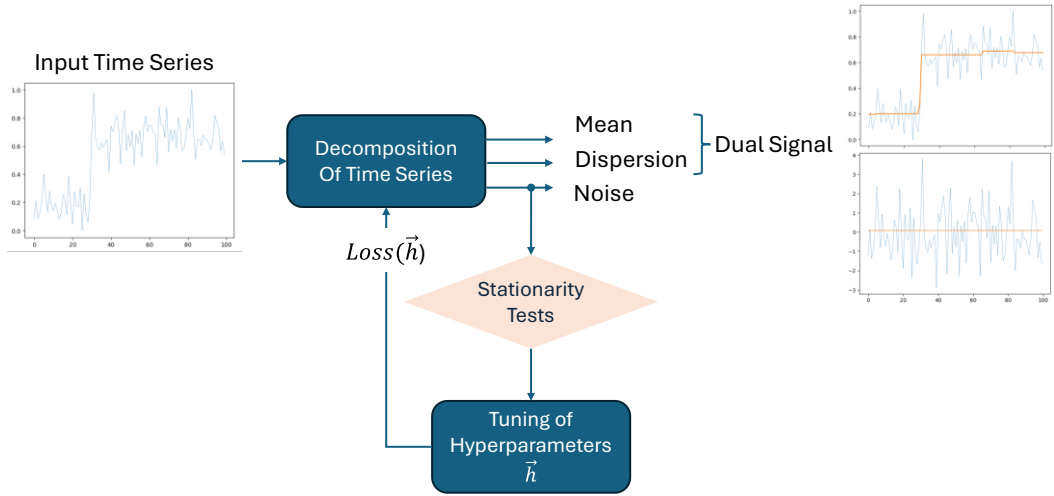

Figure 4. Architecture of the machine leaning

The vector of the loss function hyperparameters includes many elements:

$$\vec{h} = \{\alpha, \beta_M, \beta_S, \gamma_X, \gamma_S, \theta, n, p^*, \vec{I}\}$$

where $\alpha$ directs the learning process to be sequential or jointly, the next five are Lagrangian multipliers, *n* is the sample size of the SPC window, $p^*$ is a cutoff value controlling binary weights, and vector $\vec{I}$ includes hyperparameters defining different expressions such as weighting, fitting, regularization, and smoothing components.

The Lagrangian multipliers $\beta$ can have a fixed value or be estimated, such as:

$$\beta_M = C_\beta \cdot RMSE(X_t, \bar{X})/(\sqrt{T} \cdot \overline{|\Delta(X_t)|}), \tag{6}$$

where $C_\beta$ is a tunable constant, for example, $C_\beta = 25$.

Tuning of hyperparameters $\vec{h}$ to enforce stationarity of the noise can be efficiently organized by applying Desing of Experiment (DOE) techniques (Plackett and Burman, 1946; Nelder and Mead, 1965; Montgomery, 2012; Glushkovsky, 2018). Having a large number of hyperparameters, pragmatic DOE can be achieved by applying unsupervised machine learning (Glushkovsky, 2021).

## 2.4 Control of signal's smoothness

The third smoothing component in loss functions (3) and (5) control the second order variations of the decomposed signals.

As mentioned earlier, there is some overlap between regularization and smoothing. The synthetic stochastic time series that includes the linear mean trend illustrates smoothing effects by applying different regularization and smoothing components (Figure 5).

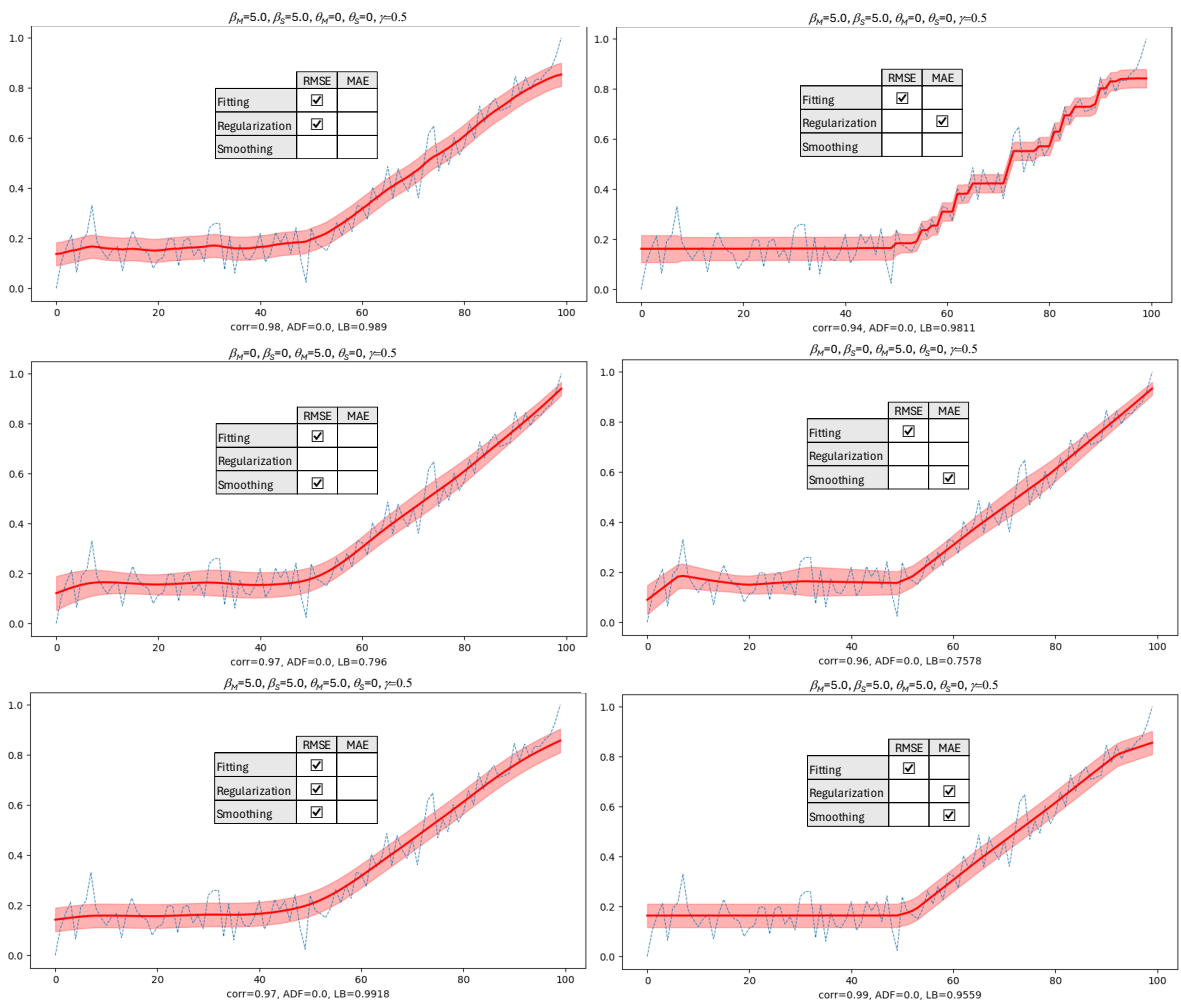

Figure 5. Illustration of the effects of regularization and smoothing components

The examples are without special effects preservations (i.e., $w_t \equiv 1$). The dashed blue line represents the original time series, the solid dark red line represents the mean signal $M_t$, and the width of the red ribbon represents the dispersion signal $S_t$ having [$M_t - S_t$, $M_t + S_t$] range.

The example includes a stochastic time series with generated Gaussian normally distributed random noise ~ $N(0,1)$ and two periods, the first half having a steady process and then the second half with the linear mean trend.

It can be observed that without the smoothing, regularization based on RMSE provides a more smoothed outcome while MAE leads to the stepped mean signal (Figure 5, top charts). It is an expected behaviour since the later option is more aligned with the Potts functional that allows for step detection. Both versions isolate stationary noise with similar ADF and LB test statistics, but the latter has a higher correlation to the introduced noise.

Decompositions applying smoothing without regularization components provide less favorable results concerning stationarity of the isolated noises (Figure 5, middle charts). Also, the learned dispersion signals $S_t$ in that case have a decreasing tendency along time that can be observed by the width of the red ribbon in both middle charts.

The best results have been achieved by applying both regularization and smoothing components (Figure 5, bottom charts). However, MAE metrics for both components provide a better outcome compared to RMSE concerning correlation between introduced and isolated noises and a sharp change from the steady period to the trend. The latter has been achieved even without applying weightings based on the statistical process control.

## 2.5    Learning of dual signal

Dual signal can be learned sequentially or jointly. Sequential learning provides independent learning of the mean signal $M_t$ first and then of the dispersion signal $S_t$. The loss functions in that case are presented in Equations (3) and (5), correspondingly.

The loss function for jointly learning can be defined as

$$Loss_{\{M,S\}} = Loss_{\{M\}} + \theta \cdot Loss_{\{S\}} \qquad (7)$$

Or

$$Loss_{\{M,S\}} = arg\ min_{\{M_t, S_t\}}(RMSE(X_t, M_t) + \beta_M \cdot E[w_t \cdot |\Delta M_t|] + \gamma_M \cdot E[w_t \cdot |\Delta^2 M_t|] +$$

$$\theta \cdot (RMSE(|X_t - M_t|, S_t) + \beta_S \cdot E[w_t \cdot |\Delta S_t|] + \gamma_S \cdot E[w_t \cdot |\Delta^2 S_t|])$$

where $\theta$ is an additional Lagrangian multiplier that controls for the balance between mean and dispersion loss components.

Both sequential and jointly learning approaches provide decomposition of the input time series (Figure 4). The decomposition process essentially corresponds to a non-linear unconstrained optimization task that minimizes the loss function and may include different set-ups, such as neural networks or direct optimization problem solving.

Simple neural networks architectures are presented in Figure 6 with customized loss functions according to Equations (3), (5) and (7). The sequential learning includes sequential layers while jointly learning includes a shared feature extractor followed by twin parallel networks that separate learning of two output signals.

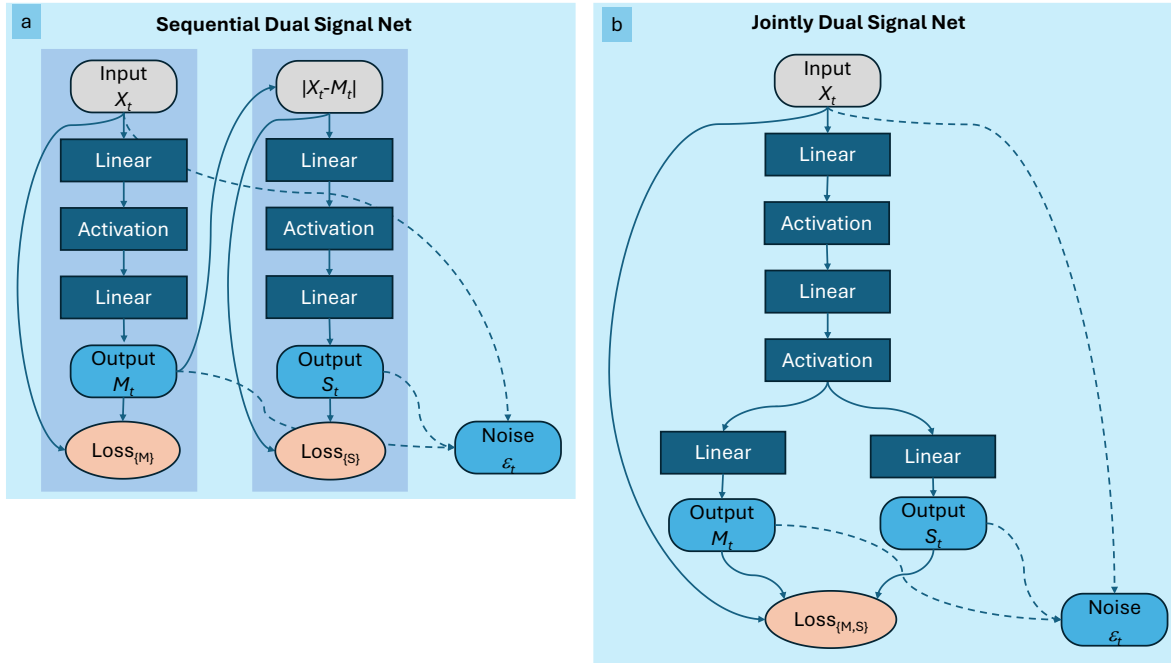

Figure 6. Examples of dual signal neural networks: (a) sequential and (b) jointly

It was shown that neural networks that jointly learn additive models can uncover complex relationships between input features and the output (Agarwal *et al*, 2020). However, the loss function (7) includes five Lagrangian multipliers that exhibit strong interactions, which significantly complicates the tuning process. Furthermore, the neural networks have an increased number of hyperparameters, such as the number of layers, the number of nodes per layer, and different activation functions. That complicates the tuning process even further.

The third neural network option with a modified twin output net has been considered. That option includes the calculation of the noise time series $\varepsilon_t = (X_t - M_t)/S_t$ as part of the network and modification of the loss function (7). The modified loss function has additional term ensuring the noise to be stationary. However, this option proved too complex due to the large number of hyperparameters requiring tuning.

Alternatively, learning can be set-up as a direct non-linear unconstrained optimization problem solving. Two methods have been explored in addition to the neural networks: (1) using the scipy.optimize package with the Broyden-Fletcher-Goldfarb-Shanno (BFGS) minimization algorithm (Nocedal and Wright, 2000) and (2) Excel Solver applying generalized a reduced gradient (GRG) non-linear method.

The standard Excel Solver can be activated in Excel as optional Add-ins. It is suitable for short time series and can run against up to 200 learned decision variables. This method allows for sequential decompositions of time series of up to 200 observations (T ≤ 200). Alternatively, it can be set to jointly learn mean and dispersions of up to 100 observations. By using Excel Solver, coding is not required since the loss functions can be calculated in Excel by applying the built-in functions.

It turns out that applying the same Lagrangian multipliers, the direct optimization and neural networks provide significantly different outputs. Therefore, dedicated tuning is required using different optimization methods. Direct optimisation is a more robust approach and does not require normalization of the input time series, but it runs significantly slower compared to the neural networks approach.

Examples of applying direct optimization (left charts) versus neural networks (right charts) are shown in Figure 7. Examples include two synthetic stochastic time series with constant mean but having heteroscedasticity effects. The top charts have a significant up shift in variance that occurred in the second half of the observation period, while the bottom charts have an increasing variance trend during the same period.

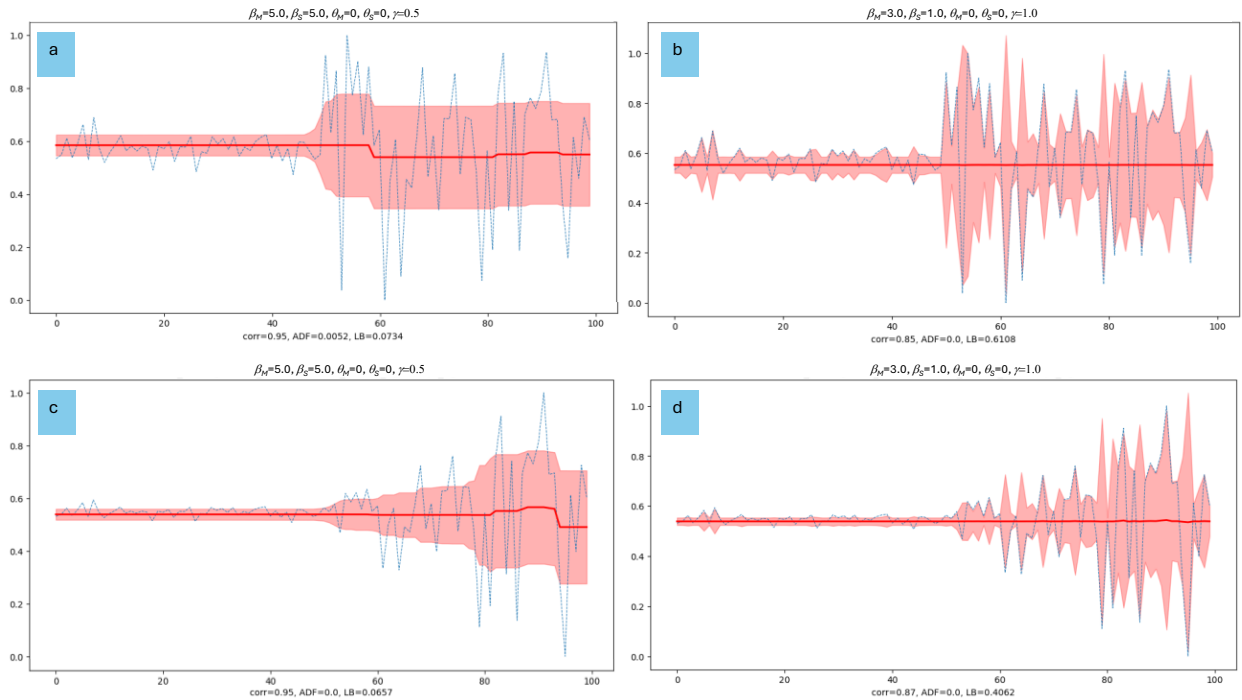

Figure 7. Decompositions of time series with shifted up (top charts) and an increasing trend (bottom charts) heteroscedasticity effects:
- the left charts (a) and (c) are decompositions applying direct optimization
- the right charts (b) and (d) are decompositions applying twin output neural networks

Jointly learning of dual signals have been used in both learning methods with switched-off smoothing components and without weightings (i.e., $w_t \equiv 1$).

It can be observed that both learning methods provide quite reasonable outputs. Thus, the augmented Dickey-Fuller and Ljung-Box tests suggest that isolated noises are quite stationary. However, the direct optimization outputs have higher correlations between introduced simulated and decomposed noises, better learning of the dispersion signals that are significantly smoother, but slightly less stable mean signals compared to the neural networks.

## 3      Examples with various effects

Let us consider two illustrative examples having various effects: (1) a synthetic time series, and (2) the real-world Canadian Consumer Price Index (CPI).

## 3.1 Synthetic time series

A synthetic time series (Figure 8, a) designed to exhibit a range of effects including outliers, shifts, trends, cycles, and periods of heteroscedasticity. These effects are superimposed on a baseline of stationary Gaussian white noise distributed as $\sim N(0,1)$. There are periods that remain unimpacted by any effects—i.e., stationary stochastic periods—and others that are influenced by single, double, or even triple effects. These affected periods are represented by horizontal bars in corresponding colors.

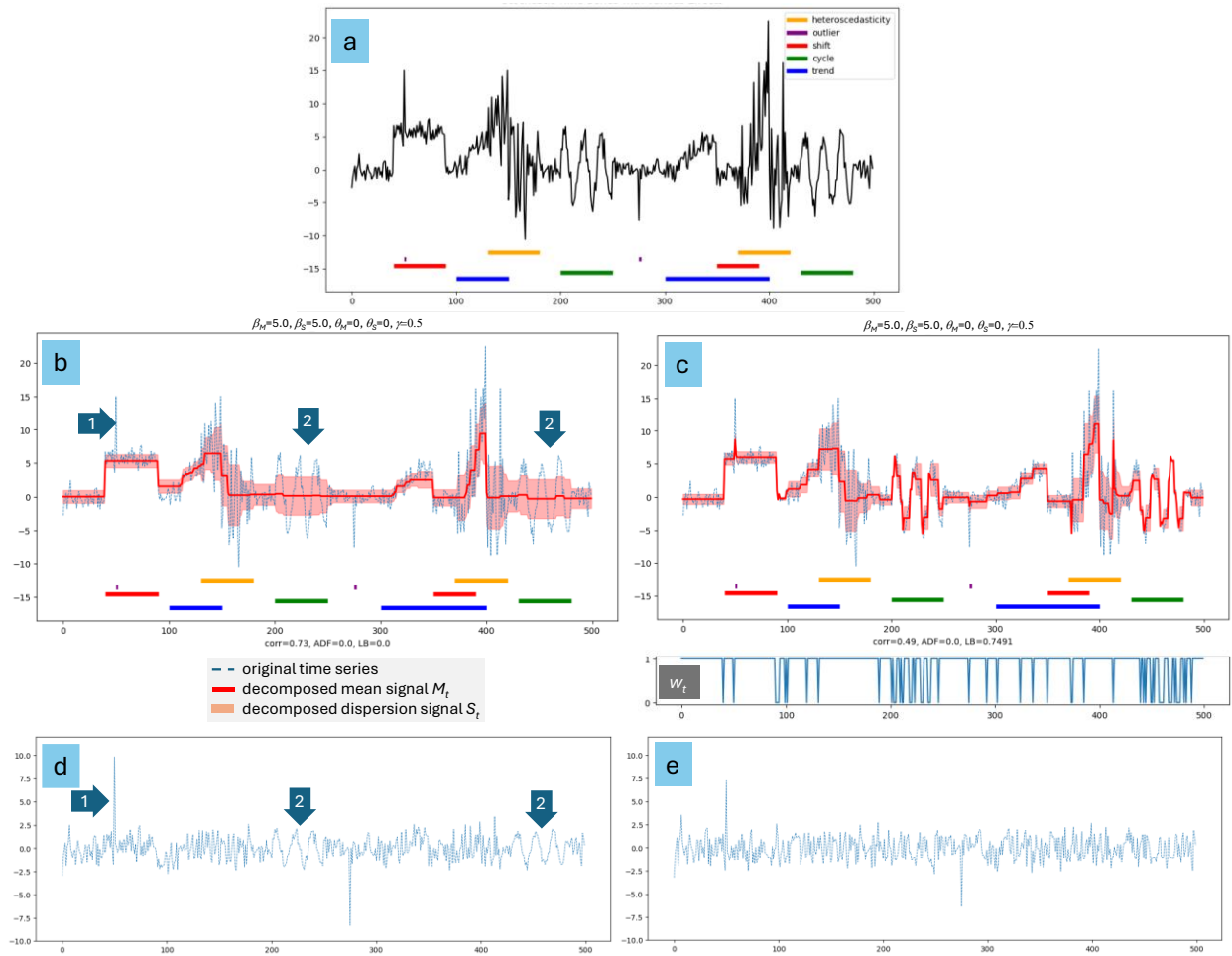

Figure 8. Illustrative example of synthetic time series with various effects (a):
- sequentially learned dual signals (b) and isolated noise (d) without preserving special patterns, i.e, $w_t \equiv 1$
- jointly learned dual signals (c) and isolated noise (e) applying binary weightings to preserve special patterns, where $w_t = 0$ corresponds to "out-of-control" state, and $w_t = 1$, otherwise.

Figures 8, (b) and (c) provide results of the dual signal learnings, first applying a sequential process without preserving special patterns, and second running a jointly direct optimization method applying binary "out-of-control" weightings. It can be observed that there are many "out-of-control" states where $w_t = 0$.

It turned out that the sequential learning ignores the outlier (marked by arrow #1) and incorrectly recognizes cycles as heteroscedasticity periods (marked by arrows #2). As a result, the isolated noise (Figure 8, d) has an outlier's spike and deterministic cycled patterns rather than being stochastic. While the augmented Dickey-Fuller test has a very low p-value suggesting stationarity, the Ljung-Box test indicates that the noise values are autocorrelated.

By contrast, jointly learning with the weightings that preserves the special patterns provided the adequate dual signal (Figure 8, c). Furthermore, isolated noise (Figure 8, e) is stationary and independently distributed according to the augmented Dickey-Fuller and Ljung-Box tests, correspondingly. It is quite an expected outcome considering that weightings prevent over smoothing especially having cycles and outliers.

Overall, the sequential learning where the mean $M_t$ and dispersion $S_t$ signals are independently fitted and tuned is faster and easier to control since it has less hyperparameters. However, it has limited abilities to discover inherent dependence between the dual signal compared to jointly learning.

### 3.2 Canadian Consumer Price Index

The real-world Canadian Consumer Price Index (CPI) shown in Figure 9, (a) provides a quarterly measure of core inflation (Bank of Canada, 2025). The applied dual signal decomposition is based on jointly learning and incorporates preservation of the special signal patterns by binary weightings $w_t$. It reveals quite stable periods mostly within the target range set by Bank of Canada as well as very volatile periods especially during the U.S. subprime mortgage collapse (2008-2009) and during and post the global COVID-19 pandemic (2020-2024).

Learned dual signal time series can be presented on the ($M_t$, $S_t$) space. This representation enhances time series analysis towards the following aspects:

- Locations of the scattered dots in Figure 9, (b) are the process states that can be analyzed by using common statistics, such as correlation or mutual information. The latter is a sounder metric considering usually non-linear dependency between signals. These statistics can be seen as metadata describing the original time series.
- Interpolated 2D density distribution estimates probability to be in a particular state on the dual signal space (Figure 9, c). Density values estimate periods (cycle time) to be in a particular state on that space. It can provide additional valuable information for exploratory analysis. Also, it allows for the identification of outliers having extremely low 2D densities.
- Trajectories on the 2D space reveals dynamic dependencies between dual signal time series. Transition from one state to another one can be presented as a directed graph (Figure 9, b).
- Interpolated vector field characterises the state-to-state transitions (Figure 9, d). It supports Markov chain formalization (Grimmett and Stirzaker, 2001), where states are presented by the points and state-to-state transitions are vectors between consequent states along the timeline. Combined with the interpolated 2D density distribution, it can be used to recurrently forecast the next location on the 2D signal space given the current position.

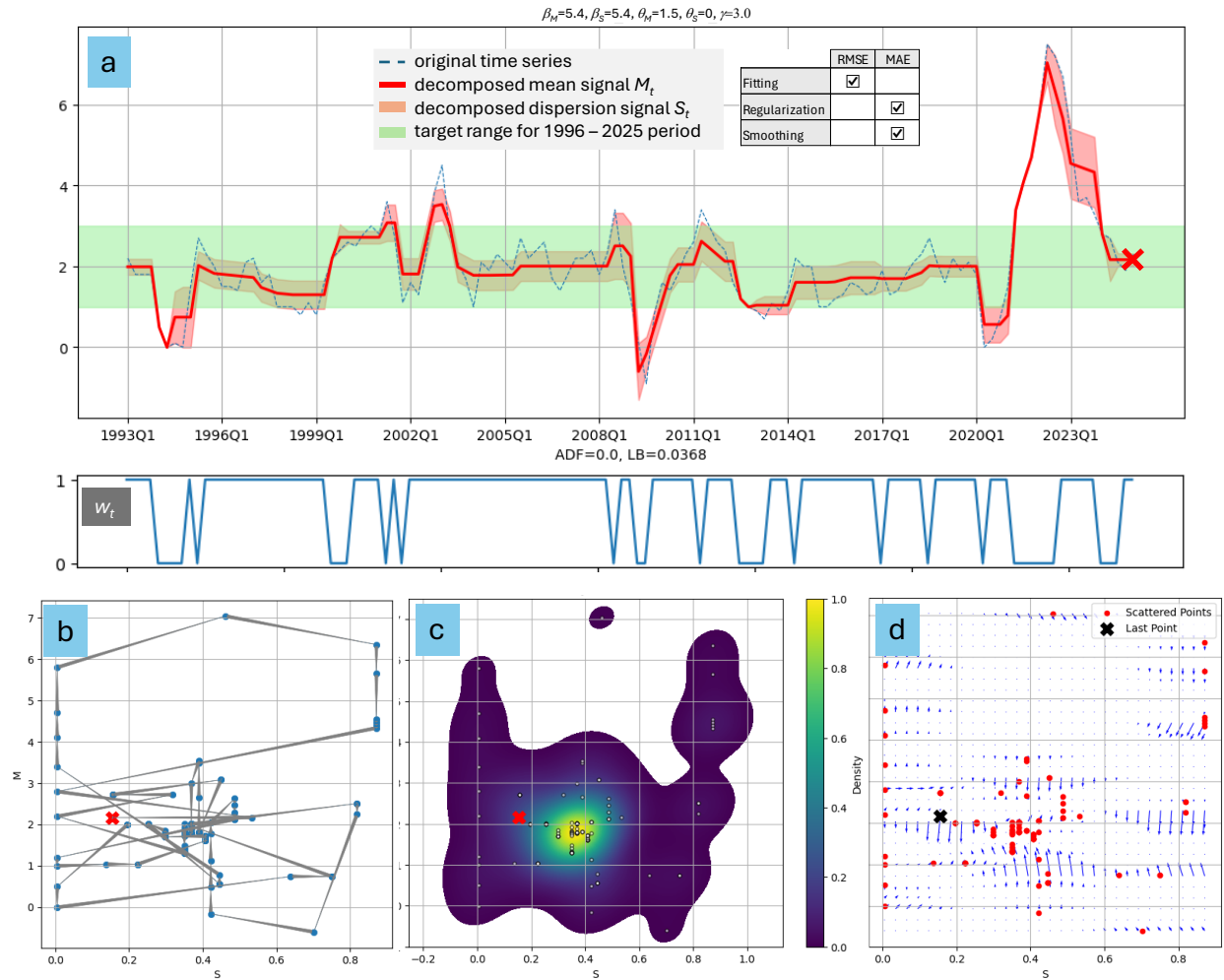

Figure 9. The real-world illustrative example:
  (a) Canadian Consumer Price Index time series
  (b) trajectory on the dual signal space, where time flow is indicated by the line thickness from the thick ends to thin ends
  (c) density plot on the dual signal space
  (d) vector field plot on the dual signal space

The mentioned above representations on the dual signal space support utilization of the well-developed machine learning algorithms where input features are coordinates of the scattered points and trajectories:

- Scatterplots can be converted into images by applying normalization and discretization towards pixels. This approach reveals the potential for utilizing Convolutional Neural Networks (CNN) (Dumoulin and Visin, 2018). Being invariant to time flow, it, nevertheless, can be used to model inherent structures of multiple time series.
- Trajectories can be formalized as graphs where adjacency matrixes reflect the state-to-state directed transitions and vertices are points on the dual signal space. This formalization can be fitted by Graph Neural Networks (GNN) (Scarselli *et al*, 2009). It allows for forecasting incorporating 2D neighboring distances, interpolated vector field, and density distribution.

The mentioned above 2D representations as input features in CNNs or GNNs can be especially helpful in analyzing cross effects in case of multiple time series.

## 4    Conclusions

The proposed machine learning approach allows for dual signal decomposition and isolation of the stationary noise. It can be applied for smoothing and/or denoising. By isolating the noise component, the learned dual signal can be analyzed further to discover trends, shifts, cycling (seasonality), or outliers. It supports a robust approach for discovering special patterns.

Dual signal learning can be performed sequentially or jointly. The former approach first independently fits the time series by the mean signal and then learns the dispersion signal while the jointly learning approach fits the dual signal simultaneously. Sequential learning is biased toward the mean decomposition while diminishing the role of the heteroskedasticity properties of the underlying time series. While requiring more computational time, jointly learning better reveals inherent dependencies between dual signal time series.

Learning can be performed by setting a direct non-linear unconstrained optimization problem or by applying neural networks. The direct optimisation is a more robust approach and does not require normalization of the input time series, but it runs significantly slower compared to the neural networks fitting. The sequential learning by neural network includes sequential layers while the jointly learning includes a shared feature extractor followed by twin parallel networks that separate learning of two output signals.

To preserve signal patterns, weighting of the regularization component of the loss function has been applied based on Statistical Process Control methodology. The assignable patterns commonly are masked in most decomposition or smoothing approaches. Thus, if the "out-of-control" probability of the time series point is high, meaning that the point represents a signal rather than common variance, then the impact of the regularization term is diminished. As the result, the outcome of that point stays close to the original time series value.

Tuning of the loss function hyperparameters has two objectives. First, to force the isolated noise into a stationary stochastic process without autocorrelation properties. Second, depending on applications, the hyperparameters can be tuned either toward representation of the states by discrete/stepped dual signals or toward smoothed output time series. On one hand, the large number of hyperparameters provides flexibility for achieving both objectives. On the other hand, it makes the process quite complex, often requiring the application of design of experiments techniques.

Decomposed mean and dispersion time series can be presented on the dual signal 2D space. It can be used to learn inherent structures, to forecast both mean and dispersion, or to analyze cross effects in case of multiple time series. The latter can be achieved by using 2D representations as input features for convolutional or graph neural networks.

**Disclaimer**

The paper represents the views of the author and do not necessarily reflect the views of the BMO Financial Group.